\documentclass[10pt,twocolumn,letterpaper,table]{article}
\usepackage[xcdraw]{xcolor}
\usepackage{cvpr}
\usepackage{times}
\usepackage{epsfig}
\usepackage{graphicx}
\usepackage{amsmath}
\usepackage{amssymb}
\usepackage{rotating}
\usepackage{subfig}

\usepackage{booktabs}

\usepackage[pagebackref=true,breaklinks=true,letterpaper=true,colorlinks,bookmarks=false]{hyperref}

\cvprfinalcopy 

\def\cvprPaperID{941} 

\begin{document}

\title{Adversarially Tuned Scene Generation}

\author{VSR Veeravasarapu$^{1}$, Constantin Rothkopf$^{2}$, Ramesh Visvanathan$^{1}$ \\
$ ^1$\textit{Center for Cognition and Computation, Dept. of Computer Science, Goethe University, Frankfurt}\\
$ ^2$\textit{Center for Cognitive Science \& Dept. of Psychology, Technical University Darmstadt}.
}

\maketitle

\begin{abstract}  
   Generalization performance of trained computer vision (CV) systems that use computer graphics (CG) generated data is not yet effective due to the concept of 'domain-shift' between virtual and real data. Although simulated data augmented with a few real-world samples has been shown to mitigate domain shift and improve transferability of trained models, guiding or bootstrapping the virtual data generation with the distributions learnt from target real world domain is desired, especially in the fields where annotating even few real images is laborious (such as semantic labeling, optical flow, and intrinsic images etc.). 
   In order to address this problem in an unsupervised manner, our work combines recent advances in CG, which aims at generating stochastic scene layouts using large collections of 3D object models, and generative adversarial training, which aims at training generative models by measuring discrepancy between generated and real data in terms of their separability in the space of a deep discriminatively-trained classifier. Our method uses iterative estimation of the posterior density of prior distributions for a generative graphical model. This is done within a rejection sampling framework. Initially, we assume uniform distributions as priors over parameters of a scene described by a generative graphical model. As iterations proceed the uniform prior distributions are updated sequentially to distributions that are closer to the unknown distributions of target data.   
   We demonstrate the utility of adversarially tuned scene generation on two real world benchmark datasets (CityScapes and CamVid) for traffic scene semantic labeling with a deep convolutional net (DeepLab). We obtained performance improvements by 2.28 and 3.14 points on the IoU metric between the DeepLab models trained on simulated sets prepared from the scene generation models before and after tuning to CityScapes and CamVid respectively.
\end{abstract}

\section{INTRODUCTION}

Recently, computer graphics (CG) generated data has been actively utilized to train and validate computer vision (CV) systems, especially, in situations where acquiring large scale data and groundtruth is costly. Examples are many pixel level prediction tasks such as semantic segmentation \cite{gaidon2016virtual,ros2016synthia,richter2016playing,shafaei2016play}, optical flow  \cite{fischer2015flownet} and intrinsic images \cite{Kong:ICCV:2015} etc. However, the performance of CV systems when they are trained only on simulated data is not as good as expected due to the issue of domain shift \cite{ros2016synthia}. 
This problem is due to the fact that the probability distribution over parameters resulting from the simulation process, $P(\Theta)$, may not match those parameters describing real-world data, $Q(\Theta)$. 
This can be caused by many factors such as deviations in lighting, camera parameters, scene geometry and many others from the true unknown underlying distributions $Q(\Theta)$.
These deviations may result in poor generalization 
of the trained CV models to the target application domains. The term used to describe this phenomenon is 'domain-shift' or 'data-shift'. 

In the classical CV literature, two alternatives to reduce 
domain-shift have been discussed: 1) Using engineered feature spaces that achieve invariance to large variations in specific attributes such as illumination or pose, and 2) learning of scene priors for the generative process that are optimized to the specific target domain. Several works designed \cite{baktashmotlagh2013unsupervised} or transferred the representations from virtual domains that are quasi invariant to domain shift, for instance, geometry or motion feature representations as well as their distributions (see for instance \cite{parameswaran2012design}). With the advent of automated feature learning architectures, recent works \cite{ros2016synthia,richter2016playing} have demonstrated that augmenting large scale simulated training data with a few labelled real-world samples can ameliorate domain shift. 
However, annotating even a few samples is expensive and laborious in many pixel level applications such as optical flow and intrinsic images. Hence, bootstrapping generative models from real-world data is often desired but difficult to achieve due to its inherent complexities in the bootstrapping process and the need for richly annotated seed data along with meta-information such as camera parameters, geographic information, etc. \cite{gaidon2016virtual}. 

Recently, advances in the field of unsupervised generative learning, i.e. \textit{Generative Adversarial Training} \cite{goodfellow2014generative}, popularly known as generative adversarial networks (GANs), propose to use unlabelled samples from a target domain to 
progressively obtain better point
estimates of parameters in generative models 
by minimizing the discrepancy between generative and target distributions in the space of a deep discriminatively-trained classifier. Here, we propose to use and evaluate the ability of this adversarial approach to tune scene priors in the context of CG based data generation. 

\begin{figure}[ht!]
\centering
\includegraphics[width=0.48\textwidth]{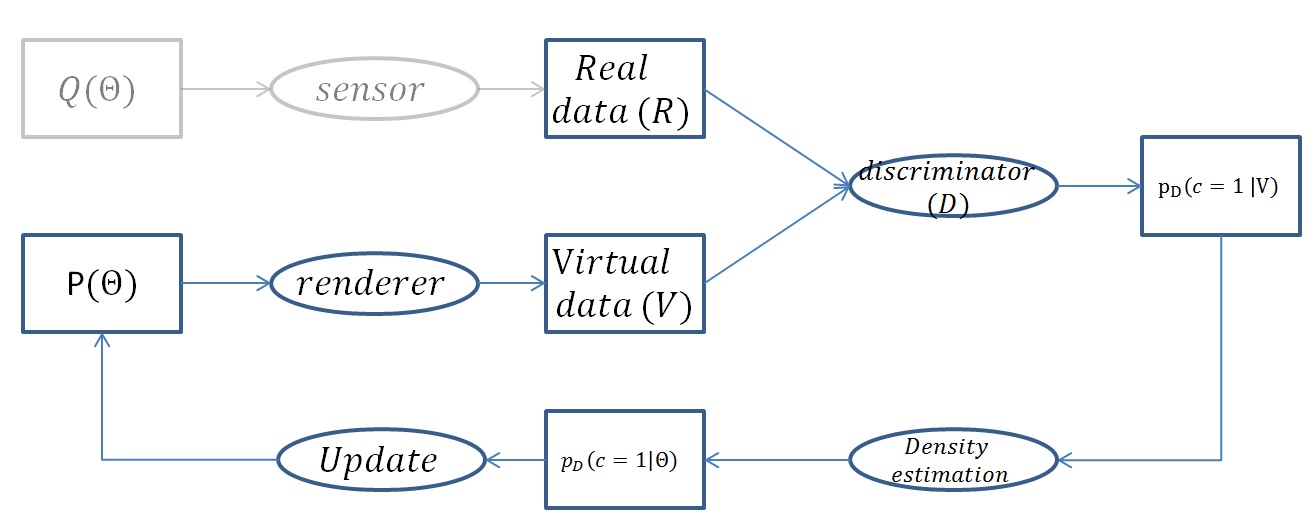}
\caption{Flow chart of adversarial tuning}
\label{fig_flow}
\end{figure}

In the traditional GAN approach neural networks are used both for the generative model and the discriminative model \cite{goodfellow2014generative,radford2015unsupervised}). Our paper focuses on the iterative estimation of the posterior density 
over parameters describing prior distributions over parameters, $P(\Theta)$, for a generative graphical model via: 1) generation of virtual samples given a starting prior, 2) estimation of conditional class probabilities of labeling a given virtual sample as real data using a discriminative classifier network $D$, 3) mapping these conditional class probabilities to estimate class conditional probabilities for labeling of data as real given the parameters of the generative model $\Theta$, and finally, 4) doing a Bayesian update to estimate the posterior density over parameters describing the prior $P(\Theta)$. This is done within a rejection sampling framework. Initially, we assume uniform distributions as priors on the parameters of the generative scene model. As iterations proceed the uniform prior distributions get updated to distributions that are closer to the unknown prior distributions of target data. 
Please see Fig \ref{fig_flow} for a schematic flow of our adversarial tuning procedure. 

More specifically, we use a parametric generative 3D scene model, $G$, which is a graphical model with scene semantics. This makes it possible to generate semantic annotations along with image data by using an off-the-shelf graphics rendering method. This model exploits existing 3D CAD models of objects and implements intra-object variations. This model is parametrized by several variables including 1) \textit{Light variables}: intensity, spectrum, position of light source, weather scattering parameters; 2) \textit{Geometry variables}: object cooccurrences, spatial alignments; 3) \textit{Camera parameters}: position and location of the camera. 

\textbf{Paper organization}: We will first review some of the related concepts and works in Section \ref{sec_background}. 
Section \ref{sec_tune} introduces our generative model and adversarial training approach to tune the model's parameters. 
Our experiments in Section \ref{sec_experiment} compare the model's properties before and after adversarial training. This includes comparing data statistics and generalization of vision systems against real world data.  Finally we conclude in Section \ref{sec_conclude} by describing future directions.

\section{BACKGROUND}\label{sec_background}
Our work builds upon several recent advances in the fields of computer graphics, which aim to automatically generate configurations of 3D objects from individual 3D CAD models of objects, and unsupervised generative learning, which aim to train generative models to a given unlabeled dataset from a target domain. Here, we summarize related work and concepts that are relevant to our work.  

\subsection{Scene Generative Models}
Automatic scene generation has been a goal both within CG and CV. The optimal spatial arrangement of randomly selected 3D CAD models according to a cost function is a well studied problem in the field of CG. Simulated annealing based optimization of scene layouts have been applied to specific domains such as the arrangement of furniture in a room. For instance, \cite{yu2011make} use a simulated annealing approach to generate furniture arrangements that obey specific feasibility constraints such as spatial relationships and visibility. Similarly, \cite{merrell2011interactive} propose an interactive indoor layout system built on top of reversible-jump MCMC (monte-carlo markov chain) that recommends different layouts by sampling from a density function that incorporates layout guidelines.  Factor potentials are used in \cite{yeh2012synthesizing} to incorporate several constraints, for example, that furniture does not overlap, that chairs face each other in seating arrangements, and that sofas are placed with their backs against a wall. 

Similarly, in the aerial image understanding literature, several spatial processes have been used to infer 3D layouts \cite{lafarge2010geometric, utasi20113}. 
Ample literature has been describing the constraints that characterize pleasing design patterns such as spatial exclusion, mutual alignment \cite{alexander1977pattern}. Inspired by these works, we view city layouts as point fields that are associated with some marks, i.e. attributes such as type, shape, scale, and orientation. Hence, we use a stochastic spatial process called a Marked Point Process, which is coupled with 3D CAD models and is used to synthesize geometric city layouts. Spatial relations and mutual alignments are encoded using Gibbs potentials between marks. 

\subsection{Graphics for Vision}
Due to the need for large scale annotated datasets, e.g. in the automotive setting, several attempts have been utilizing existing CAD city models \cite{ros2016synthia}, racing games \cite{richter2016playing, shafaei2016play} or probabilistic scene models for annotated data generation, but naturalistic scenes have even been used to investigate properties of the human visual system \cite{rothkopf2009learning}.
In the context of pedestrian detection, some work \cite{vazquez2014virtual} demonstrated domain adaptation methods by exploring several ways of combining a few real world pedestrian samples to many synthetic samples from H-life game environments. In \cite{gaidon2016virtual}, the authors introduced a fully annotated synthetic video dataset, based on a virtual cloning method that takes richly annotated video as input seed. More recently, several independent research groups \cite{ros2016synthia,richter2016playing,shafaei2016play} demonstrated that augmenting a large collection of virtual samples 
with few labelled real-world samples
could improve domain-shift. In our work, we address the question of how far one can go without the need of labelled real world samples. We use unlabelled data from a target domain and estimate the scene prior distributions of the generative model whose samples are adversary to the classifier.

\begin{figure*}
    \centering
    \includegraphics[width=0.43\textwidth, height=4.0cm]{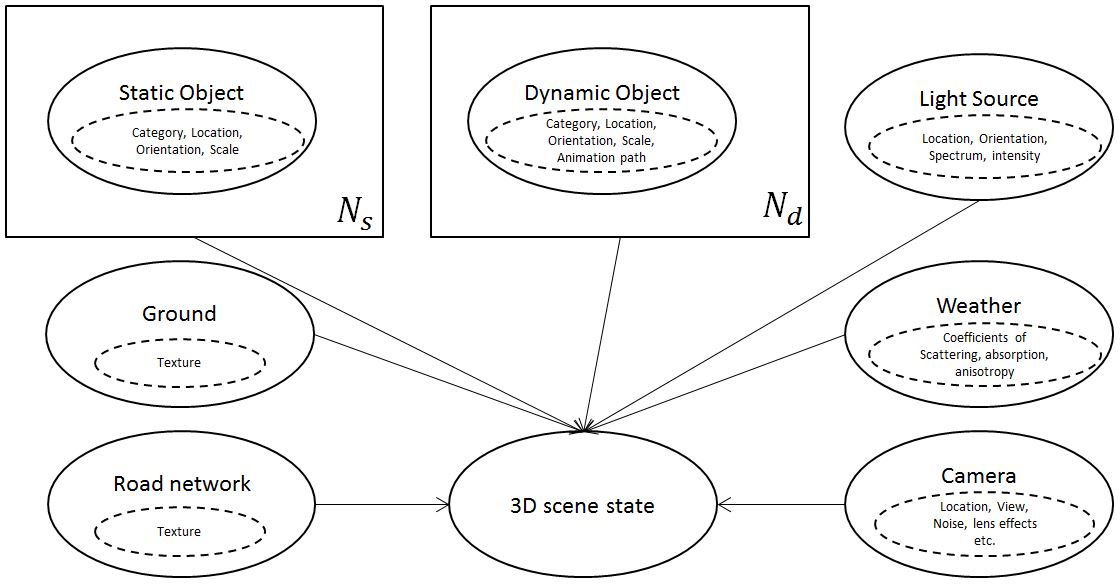}
    \includegraphics[width=0.43\textwidth, height=4.0cm]{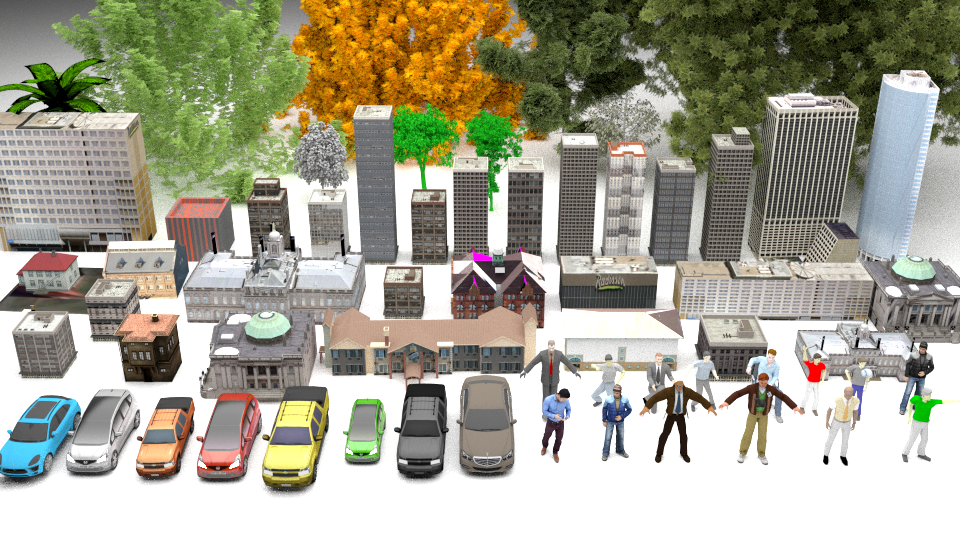}
    \caption{Graphical representation of the scene generative model and illustration of 3D CAD object models used in this work.}
    \label{fig_pgm}
\end{figure*}

\begin{figure*}[ht!]
\centering
\subfloat[RGB image]{\includegraphics[width=0.19\textwidth, height=2cm]{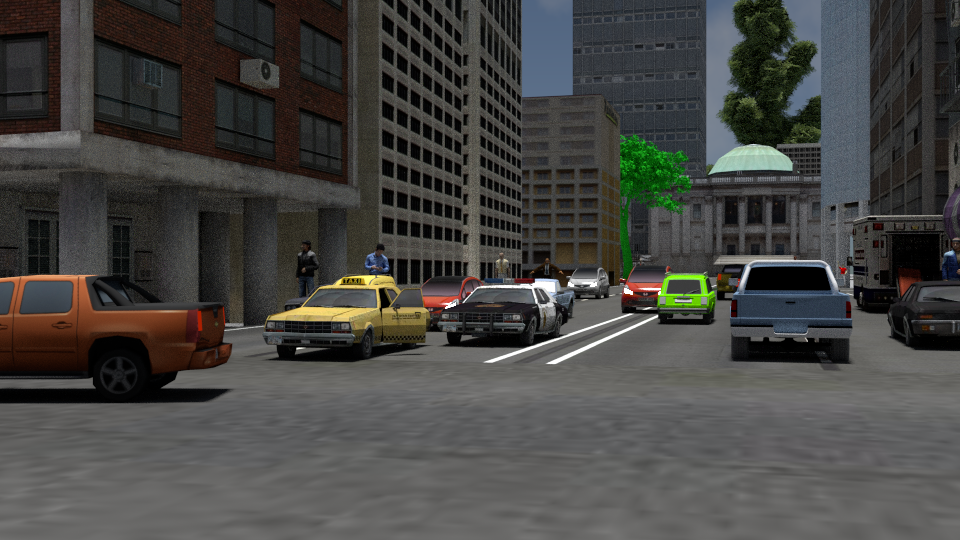}}
\subfloat[Semantic labels]{\includegraphics[width=0.19\textwidth, height=2cm]{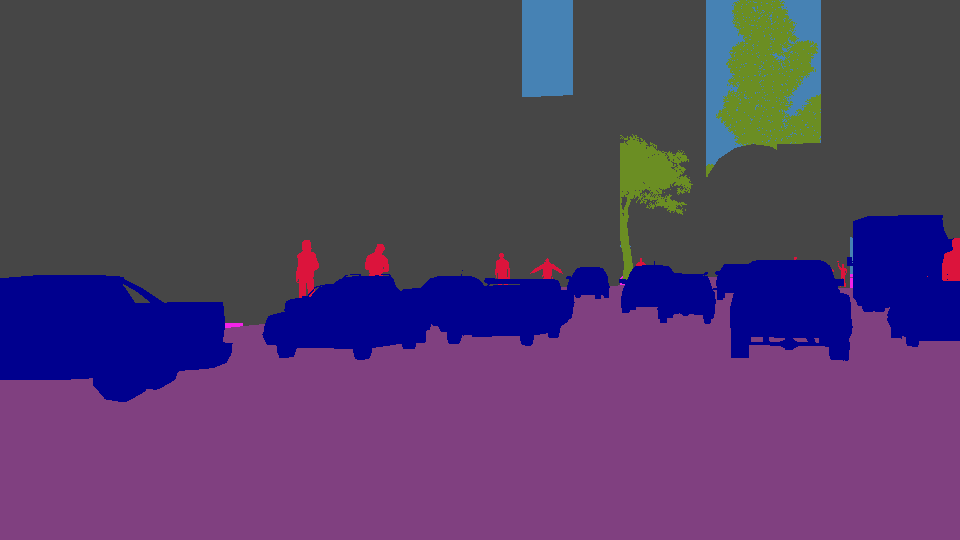}}
\subfloat[Depth]{\includegraphics[width=0.19\textwidth, height=2cm]{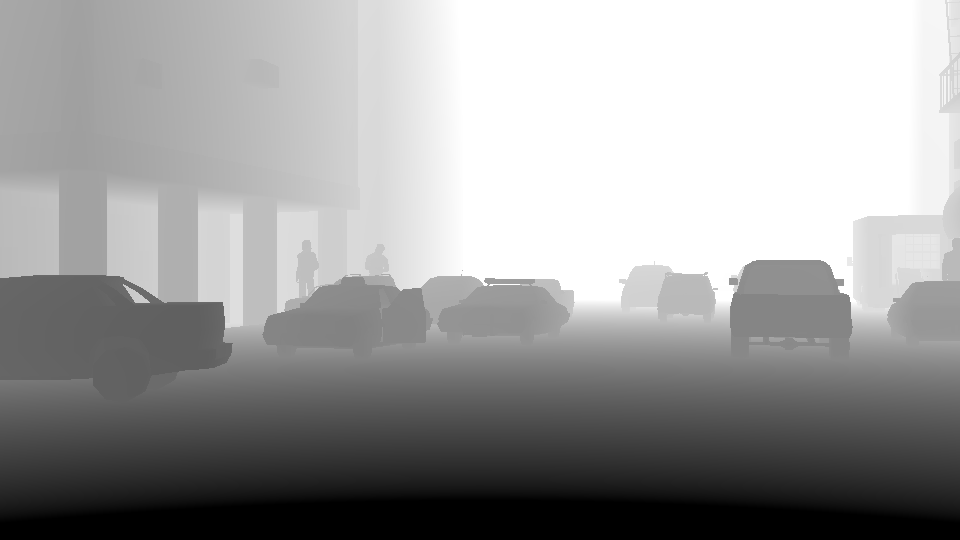}}
\subfloat[Surface normals]{\includegraphics[width=0.19\textwidth, height=2cm]{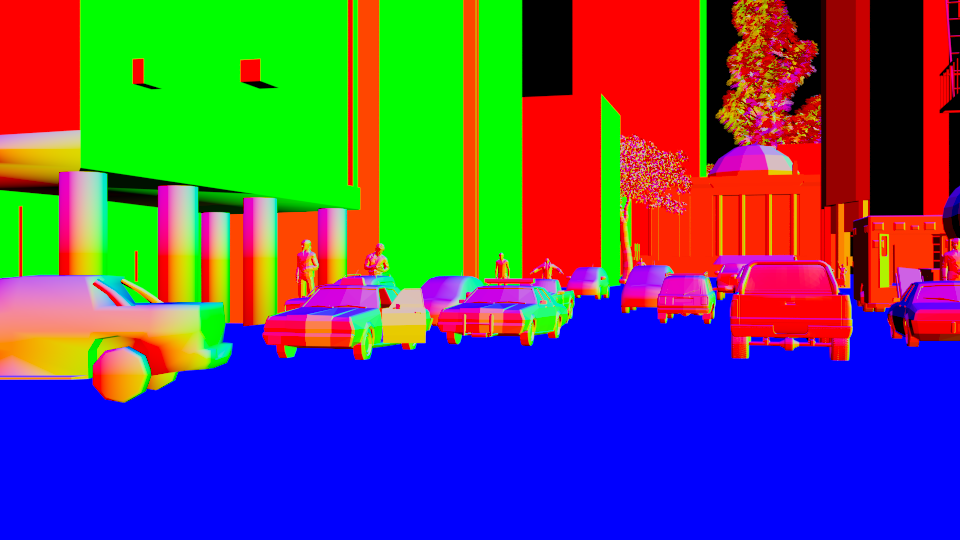}}
\subfloat[Diffuse reflections]{\includegraphics[width=0.19\textwidth, height=2cm]{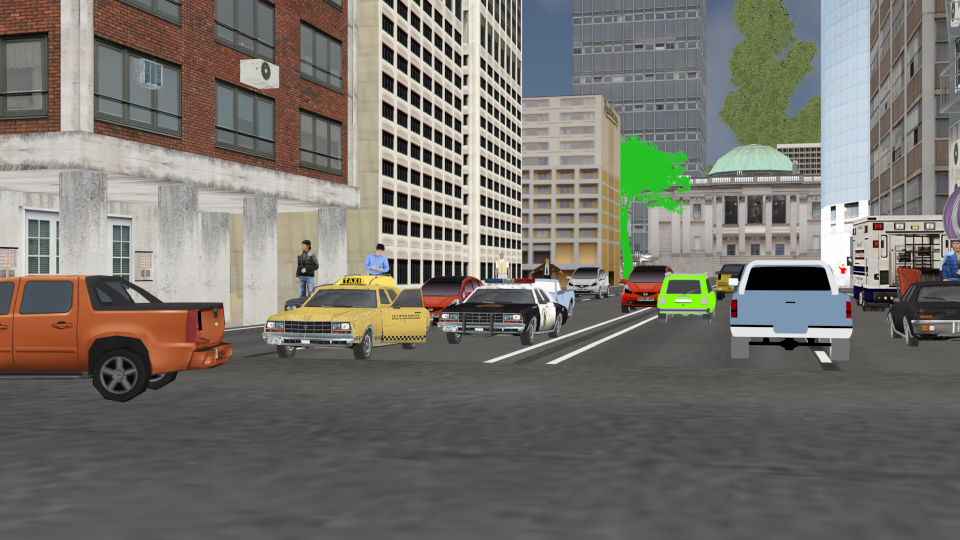}}
\caption{A rendered image sample together with corresponding pixel-level annotations.}
\label{fig_annot}
\end{figure*} 

\section{APPROACH} \label{sec_tune}
Our approach to tuning a generative model to given target-data is shown in Fig \ref{fig_flow}. We summarize the key steps below:
\begin{itemize}
\item{} The generative model $G$ has a set of parameters $\Theta$ related to different scene attributes such as geometry and photometry. 
\item{} A renderer takes these parameters sampled from the distributions $P(\Theta)$ and outputs image data $V$. 
\item{} The discriminator $D$, a standard convolutional network, is trained using gradient descent to classify data originating 
from the target domain $T$ and $V$ as being either real or generated. $D$ outputs a scalar probability, which is trained to be high if the input was real and low if the data were generated from $G$. 
\item{} The probabilities  $P(c=1|v,\Theta)$ for all simulated samples $v \in V$ are used to estimate the likelihood $P(v = real|\Theta)$. 
\item{} This is then used to update our prior distributions, which will be used in the next iteration as $P(\Theta)$. 
\end{itemize}
We now describe the details of the components used in this process. 

\subsection{Probabilistic Scene Generation}
Probabilistic scene models deal with several attributes for a scene that are relevant for the target domain. One can divide these attributes into 1) geometry, 2) photometry and 3) dynamics. However, we skip the modeling of scene dynamics in this work as we only consider static images and also aim to use publicly available large scale 3D CAD repositories such as Google's sketchup 3D warehouse. Hence, in our generative scene model we consider modeling scene layouts with CAD models and photometric parameters. 

\textbf{Scene geometry}: We designed a 3D scene geometry layout model that is based on Marked Poisson Processes coupled with 3D CAD object models. It considers objects as points in a world coordinate system and their attributes, such as object class, position, orientation, and scale as marks associated with them. These points are sampled from a probabilistic point process and the marks are sampled from another set of conditional distributions such as distributions on bounding box sizes, orientations given object type, etc. 3D CAD models are randomly imported from our collection with a few samples shown in Fig \ref{fig_pgm}, and placed in  sampled scene layouts. The camera is linked to a random car with a height that is uniformly distributed around a mean height of $1.5 \pm 0.5 m$. 

In sampling from the world models
one can assume statistical independence between marks of the point process for simplicity. Such scene states are likely to generate objects with spatial overlaps, which are physically improbable. Hence, some inter-dependencies between marks such as spatial non-overlap, cooccurrence, and coherence among instances of object classes are incorporated with the help of Gibbs potentials. In such cases, the resulting point process is called a Poisson process \cite{lafarge2010geometric} and the density of object layouts is formulated using the Gibbs equation: $\pi(o) = \frac{e^{-E(o)}}{\int_{O} e^{-E(o)}}$, where $E(o)$ introduces prior knowledge on the object layouts by taking into account pairwise interactions between the objects $o$. This allows encoding strong structural information by defining complex and specific interactions such as interconnections or mutual alignments of objects \cite{lafarge2010geometric,tournaire2007rectangular}. 
However, due to computational complexities such constraints results in extended computational times in sampling scene states. 
To avoid these problems, we limit the interactions to the essential ones for obtaining a general model of the non-overlapping objects and 
constraining road angles. 
Strong structural information can then be introduced in a subsequent step by developing post-processing in order to connect objects. This can be expressed using the term $E(o) = \sum_{o_i, o_j \in O} (e^{k L(o_i, o_j)} - 1)$, where $L(o_i, o_j)$ takes on values in the interval $[0, 1]$ and quantifies the relative mutual overlap between objects $o_i$ and $o_j$, and $k$ is a large positive real value (in our experiments $k=1000$), which strongly penalizes large overlaps. For small overlaps between two objects this prior will only weakly penalize the global energy. But if the overlapping is high, this prior will act as a hard constraint, strongly affecting overall energy.

\textbf{Scene photometry}: In addition to the above geometry parameters, we also model 1) the light source sun and its extrinsic (position and orientation) as well as intrinsic parameters (intensity and color spectrum), 2) weather scattering parameters (particle density and scattering coefficient), 3) camera extrinsic parameters such as orientation and field-of-view. These models are implemented through the use of python scripting interface to an open source graphics platform, \textit{BLENDER} \cite{blender}. A Monte-Carlo path tracer is used to render the scenes as images along with annotations, if required. Please see the supplementary material for details. A schematic graphical model is as shown in Fig \ref{fig_pgm} along with a few samples of CAD object models used in this work.

\subsection{Initialization}
As shown in Fig \ref{fig_pgm}, our generative model is a physics-based parametric model whose inputs are a set of scene variables $\Theta$ such as lighting, weather, geometry and camera parameters. We assume that all these parameters are statistically independent of each other, which provides the least expensive option for modeling and sampling. One can model dependencies using distributions on these parameters based on an expert's knowledge for a target domain or based on additional knowledge such as atmospheric optics, geographic and demographic studies.  However, in the absence of priors, we use uniform distributions in their permissible ranges. For instance, the light source's intensity in \textit{BLENDER} is modeled as $uniform(low=0, high=6)$, where an intensity level of $0$ can correspond to night while $6$ corresponds to lighting at noon. 
With these settings our model was able to render physically plausible and visually realistic images. 
This scene model was used in our previous work, which is provided as supplementary material. Performance of a vision model trained to perform semantic segmentation on simulated data was quite good on real-world data. Yet, data-shift was observed due to deviations between the scene generation statistics and the target real-world domain. Hence, in the present work, we focus on the task of matching generative statistics to those of real-world target data such as for instance CityScapes \cite{cordts2016cityscapes}.  Some samples rendered in this initial setting are shown in Fig \ref{fig_vinit}. 

\subsection{Sampling and Rendering}
Although sampling from $P(\Theta)$ is easy initially, it eventually becomes harder as $P$ gets updated iteratively through Bayesian updates: $P(\Theta) \leftarrow P(\Theta) p(.|\Theta)$. The reason is that we do not have conjugate relationships between the classifier's probabilities and $P(.)$. Hence, these intermediate probabbility functions lose their \textit{easy-to-sample-from} structure. Hence, we use a rejection sampling scheme to sample from $P$ due to its scalability. In general, an open issue in the use of rejection sampling schemes is to come up with an optimal scaling factor $M$, which results in a proposal distribution that is an envelope to the complicated distribution that we want to sample from. This issue does not arise in our case as our initial uniform distributions of $P$ can behave as envelopes for all intermediate $P$s, if they are not re-normalized. However, this ends up increase the probability of rejecting many samples and therefore generating samples becomes computationally progressively more expensive with the number of iterations. We solve this issue by normalizing intermediate probability tables with their respective maximum values. 
Corresponding labels are obtained through annotation shaders, which
we implemented in Blender. An image sample with corresponding labels are shown in Fig \ref{fig_annot}. The details about our rendering choices and their impact on the semantic segmentation results can be found in the supplementary material. 

\subsection{Adversarial Training}
In a GAN setting, the generator is supplemented with a discriminator $D$, which is trained to classify samples as real versus generated. In simple terms, the output $c$ of the discriminator should be one for a real image and zero for a generated image. One can select any off-the-shelf classifier as $D$. However, the choice of $D$ plays a critical role as it measures dissimilarity between $P$ and $Q$ in the feature space that $D$ is based on. 
Here we use AlexNet, a 5 layer convolutional neural net, as $D$ to learn the feature space automatically as in conventional GANs. Standard stochastic gradient descent with backpropagation is used to train this net. 

\textbf{Training $\textbf{D}$}: All images are resized to a common resolution of $223 X 223$, which is the default input size of AlexNet's implementation in Tensorflow. This is done to speed-up the training process and save memory. However, this has the disadvantage of missing the details of smaller objects of some pedestrians and vehicles. All real images in $T$ are labelled as 1, while simulated data is labeled as 0. Data augmentation techniques such as random cropping, left-right flipping, random brightness and contrast modifiers are applied, too, including per-image whitening. 10000 epochs are used to train the classifier.  

\textbf{Tuning $\textbf{G}$}: We now estimate the quantity $P(c=1|\Theta)$ from the classification probabilities, i.e. the softmax outputs of $D$ for all virtual samples in $V$. This is estimated using weighted Gaussian kernel density estimation (KDE). Using the classifier outputs $p(c=1|v)$ as weights we obtain:
\begin{equation}
 P(c=1 | \Theta) = \sum_{v \in V} P_d(c=1|v) K_g(\Theta_v, h)
\end{equation}
where $K_g$ a Gaussian kernel with bandwidth $h$. In our experiments, we use $h=0.1$. We explored the use of automated bandwidth selection methods  but in our experiments a default setting seemed to perform adequately.  
This KDE estimate represents the likelihood of $G$ generating samples similar to $T$ for given values of $\Theta$. 
In a Bayesian setting, this can be used to update our prior beliefs about $P(\Theta)$ iteratively as:
\begin{equation}
P^{(i+1)}(\Theta) \gets P^{(i)}(c=1|\Theta)P^{(i)}(\Theta)
\end{equation}
After a number of iterations, if $G$ and $D$ have enough capacity, they will reach a point at which both cannot improve because $P(\Theta) \rightarrow Q(\Theta)$. In the limit, the discriminator is unable to differentiate between the two distributions and becomes a random classifier, i.e. $p(c)=0.5$. However, we fix the maximum number of updates on $G$ to 6 in the following experiments.


\begin{sidewaysfigure*}
\centering
\rotatebox{90}{\ \ \ \ \ \ \ \ \  \textbf{$V_{init}$}}
\subfloat[Histogram of $V_{init}$\label{fig_vinit_heights}]{\includegraphics[width=0.18\textwidth, height=2.5cm]{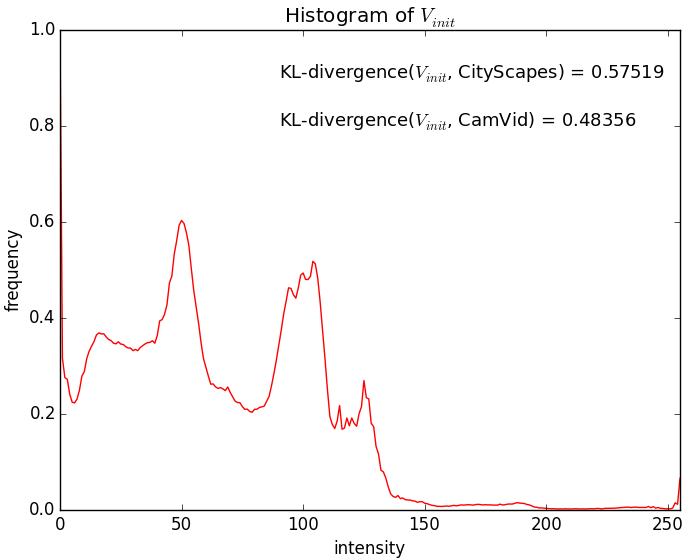}}
\subfloat[A few samples of $V_{init}$ sampled from the model before tuning\label{fig_vinit}]{\includegraphics[width=0.18\textwidth, height=2.5cm]{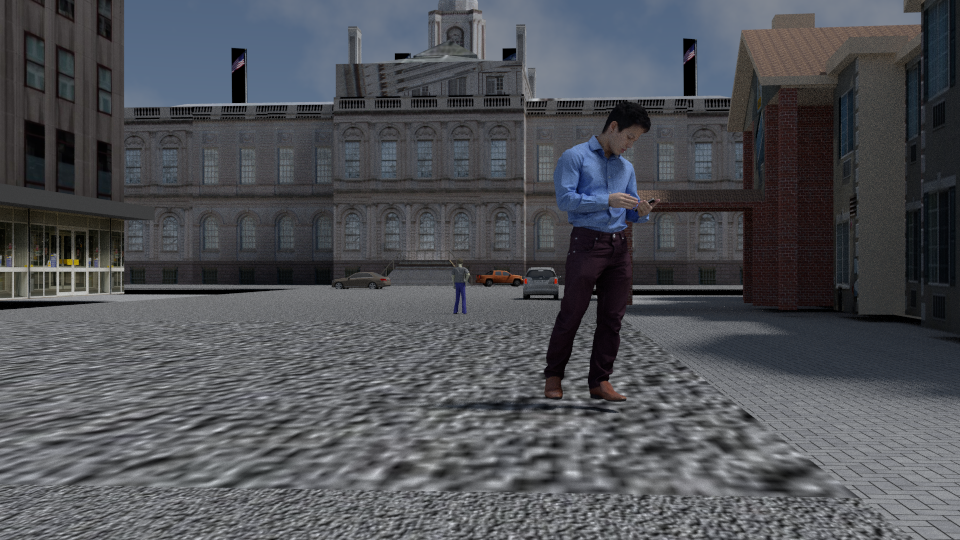}
\includegraphics[width=0.18\textwidth, height=2.5cm]{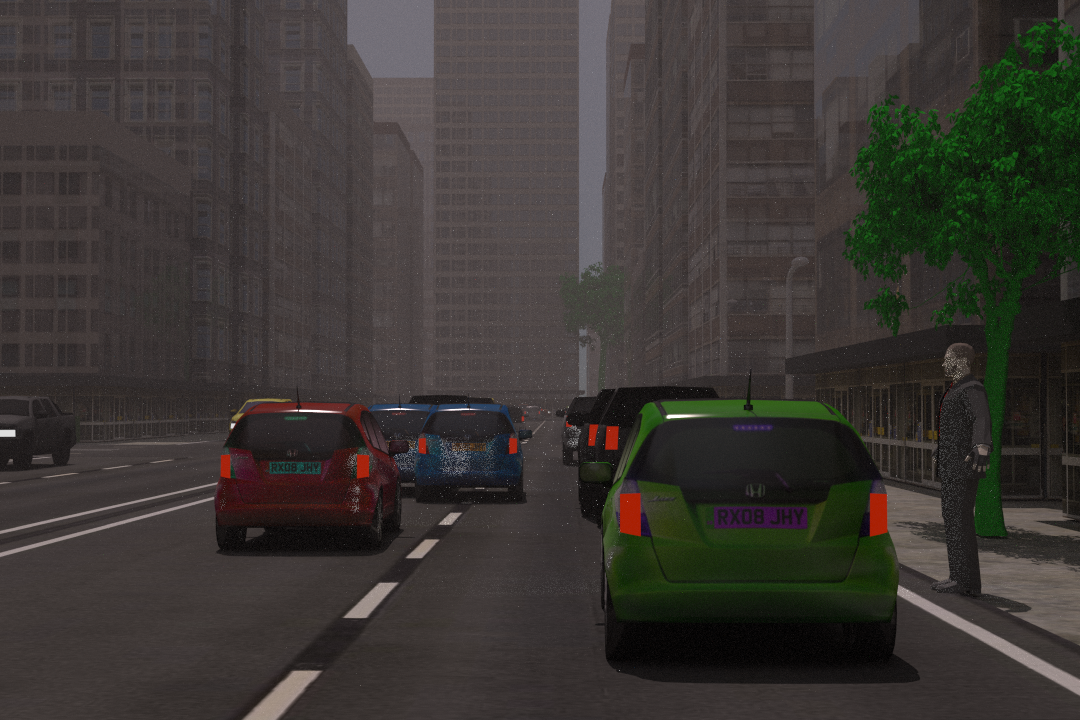}
\includegraphics[width=0.18\textwidth, height=2.5cm]{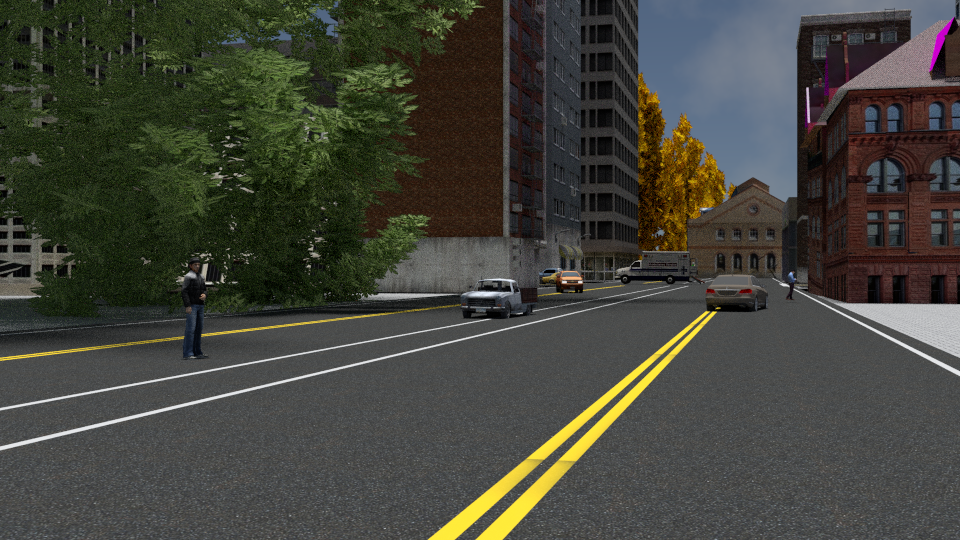}}
\subfloat[Pixel-proportions/class]{\includegraphics[width=0.18\textwidth, height=2.5cm]{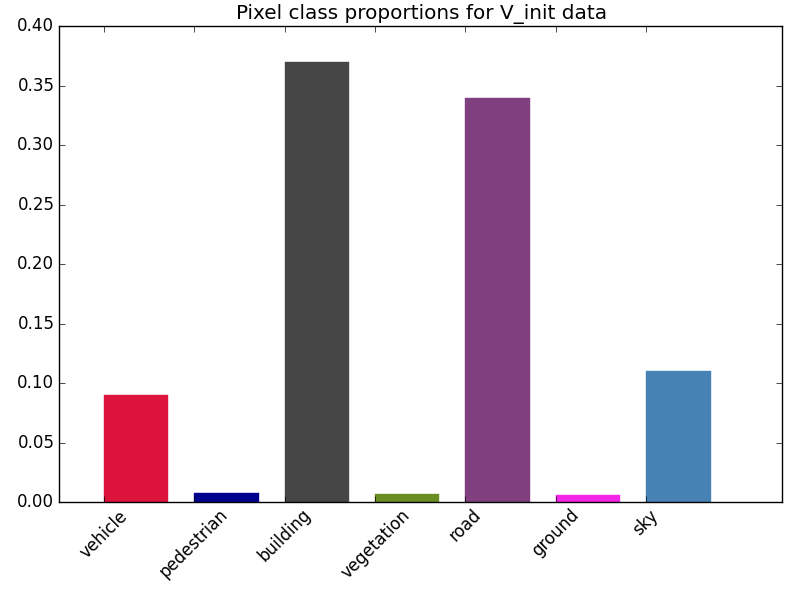}}

\rotatebox{90}{\ \ \textbf{$Cityscapes\ data$}}
\subfloat[Histogram of CityScapes\label{fig_hist_cityscapes}]{\includegraphics[width=0.18\textwidth, height=2.5cm]{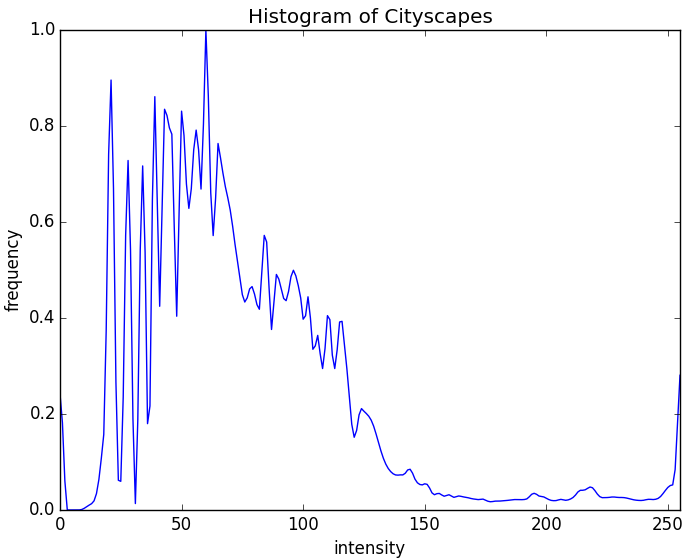}}
\subfloat[A few samples from CityScapes data]{\includegraphics[width=0.18\textwidth, height=2.5cm]{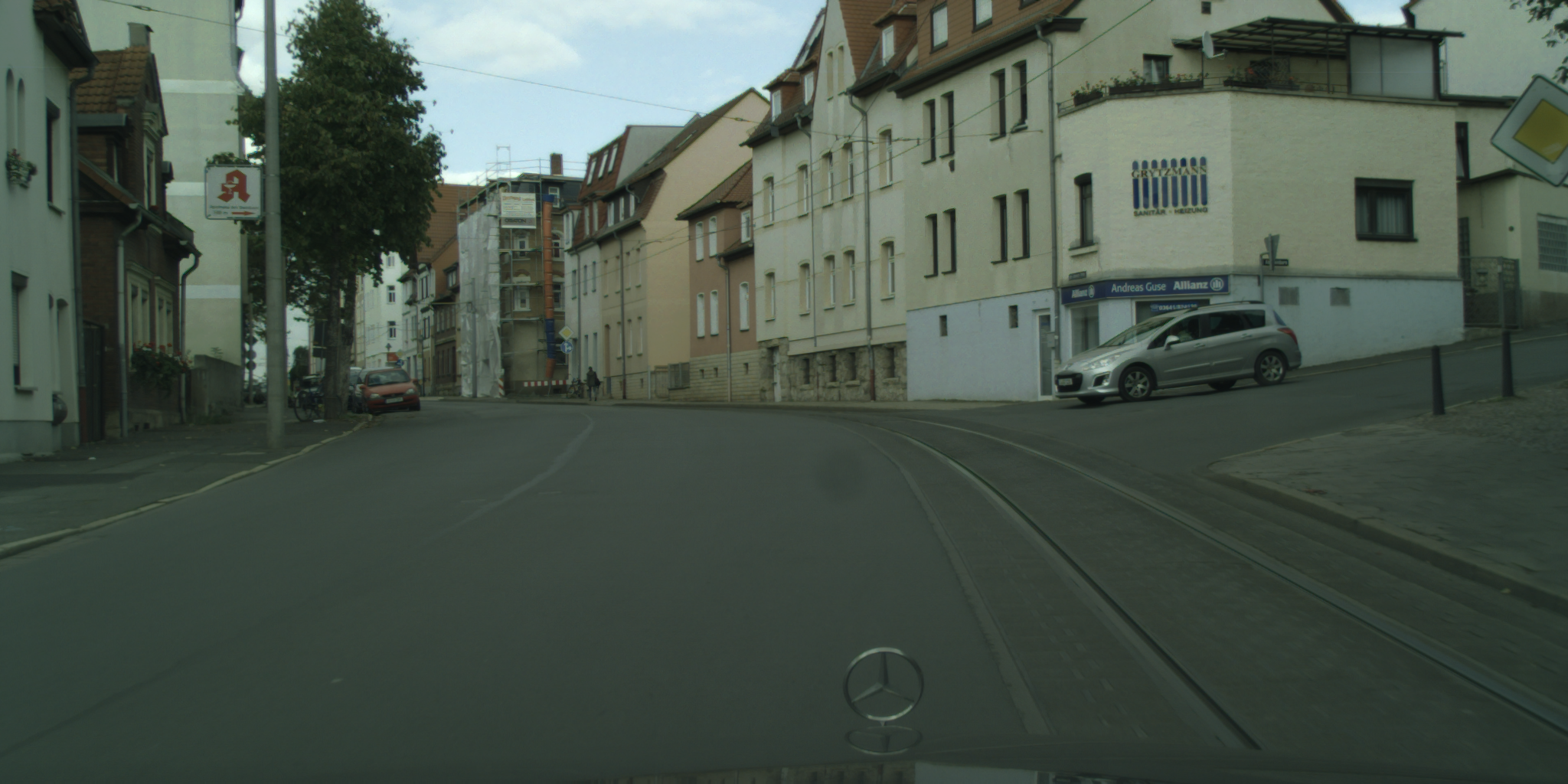}
\includegraphics[width=0.18\textwidth, height=2.5cm]{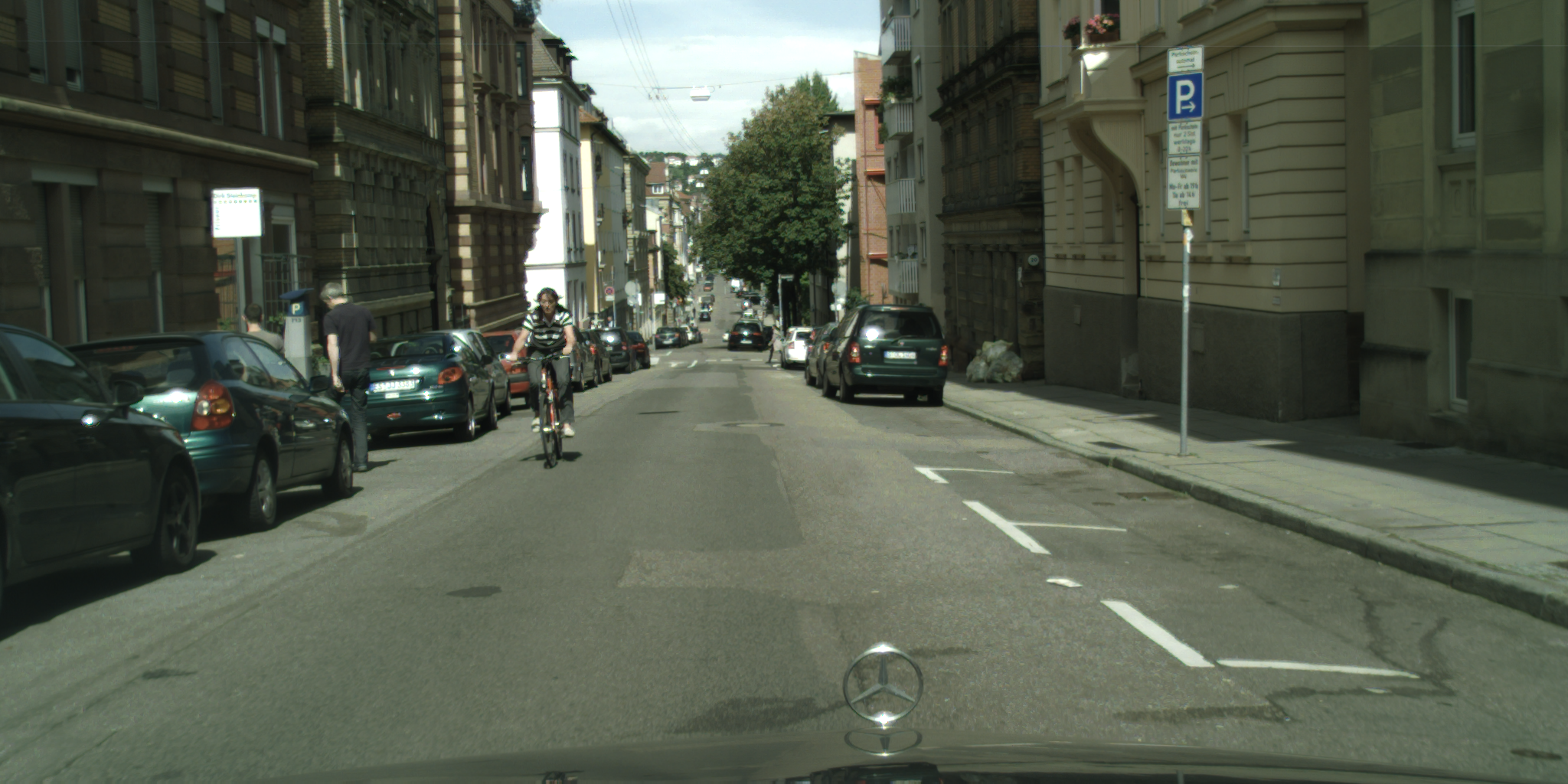}
\includegraphics[width=0.18\textwidth, height=2.5cm]{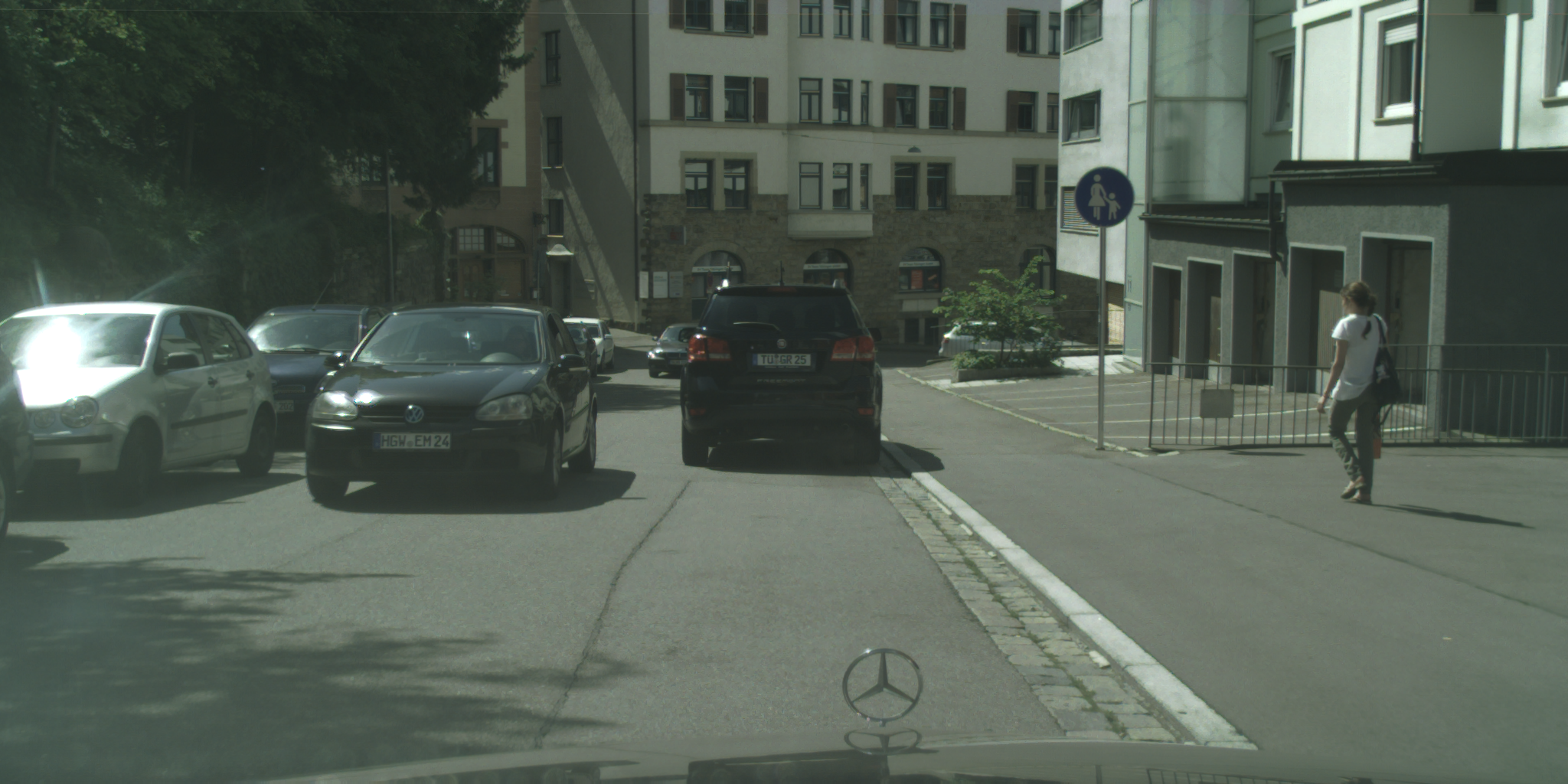}}
\subfloat[Pixel-proportions/class \label{fig_label_cityscapes}]{\includegraphics[width=0.18\textwidth, height=2.5cm]{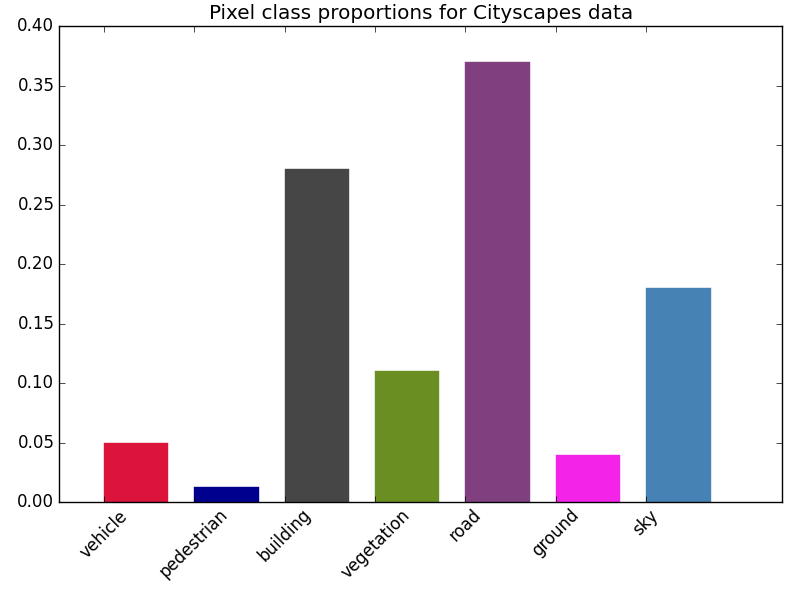}}

\rotatebox{90}{\ \ \ \ \ \textbf{$V_{cityscapes}$}}
\subfloat[Histogram of $V_{cityscapes}$\label{fig_hist_vcityscapes}]{\includegraphics[width=0.18\textwidth, height=2.5cm]{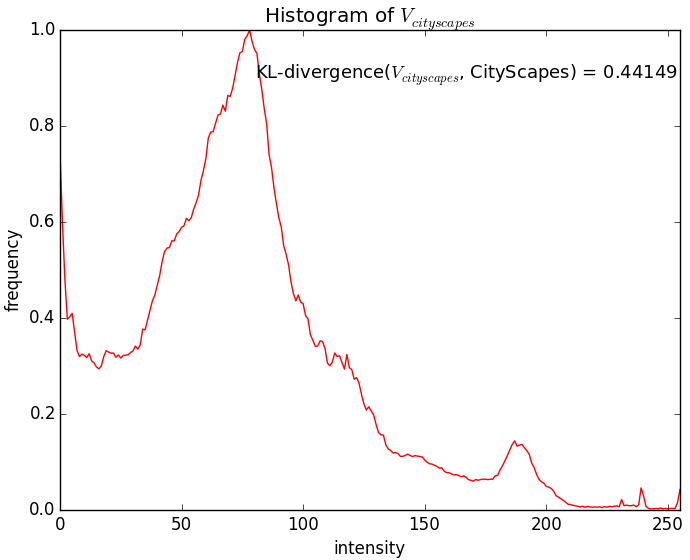}}
\subfloat[A few samples of $V_{cityscapes}$ sampled from the model after tuning \label{fig_vcityscapes}]{\includegraphics[width=0.18\textwidth, height=2.5cm]{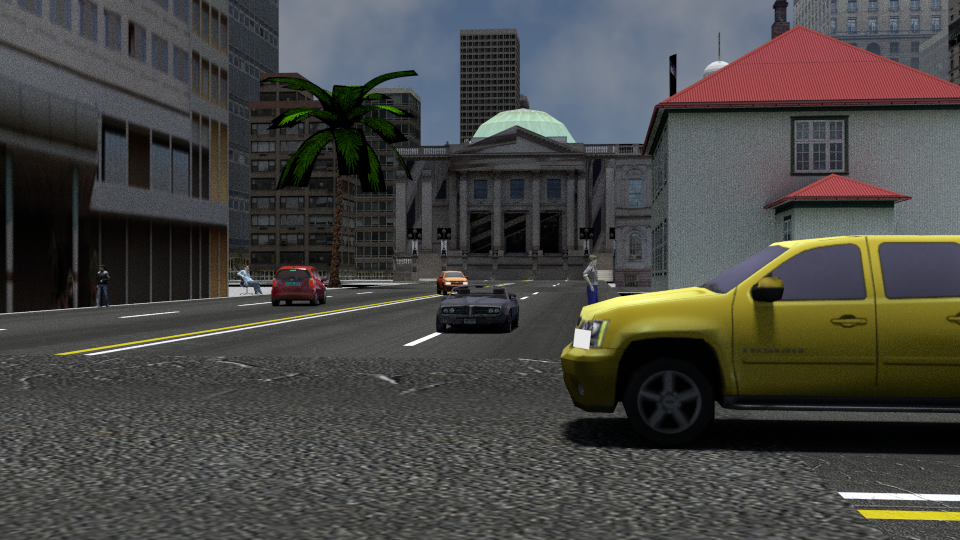}
\includegraphics[width=0.18\textwidth, height=2.5cm]{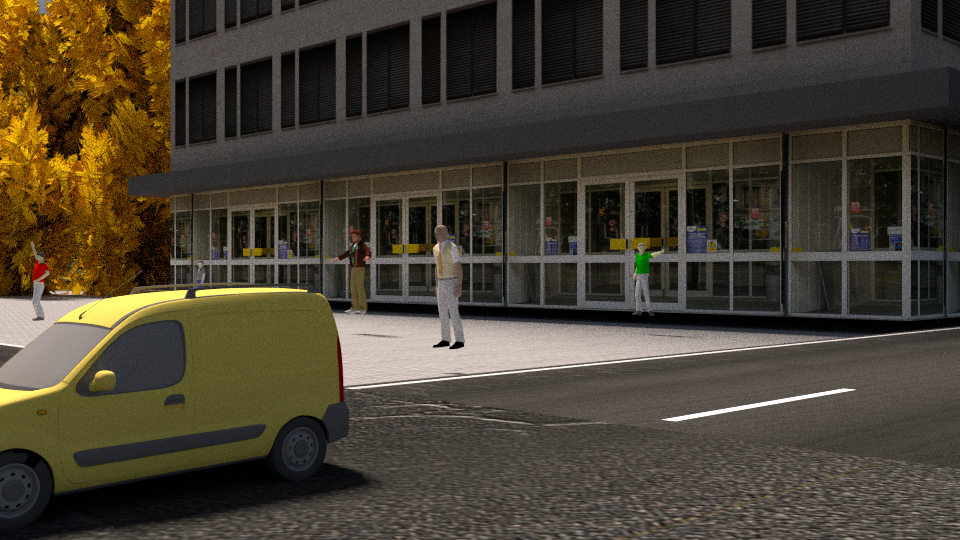}
\includegraphics[width=0.18\textwidth, height=2.5cm]{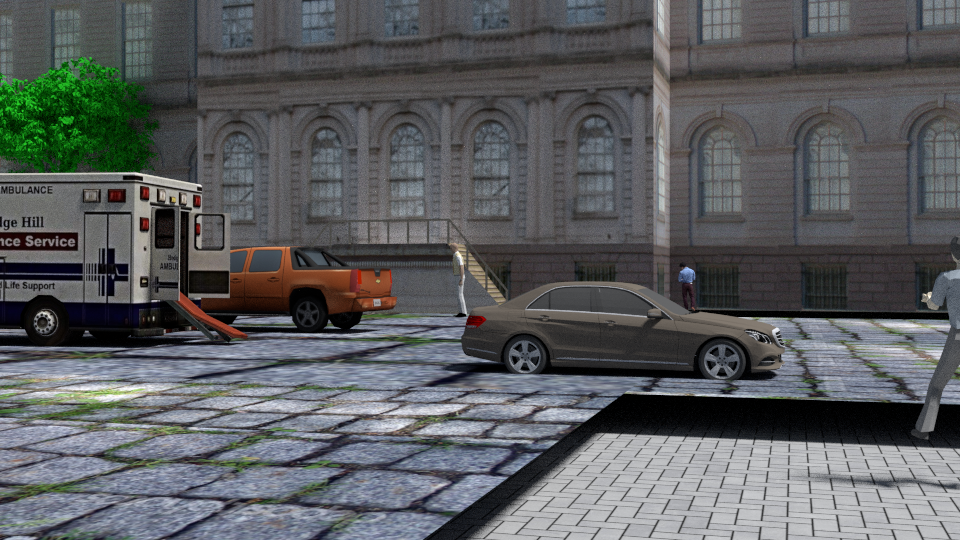}}
\subfloat[Pixel-proportions/class]{\includegraphics[width=0.18\textwidth, height=2.5cm]{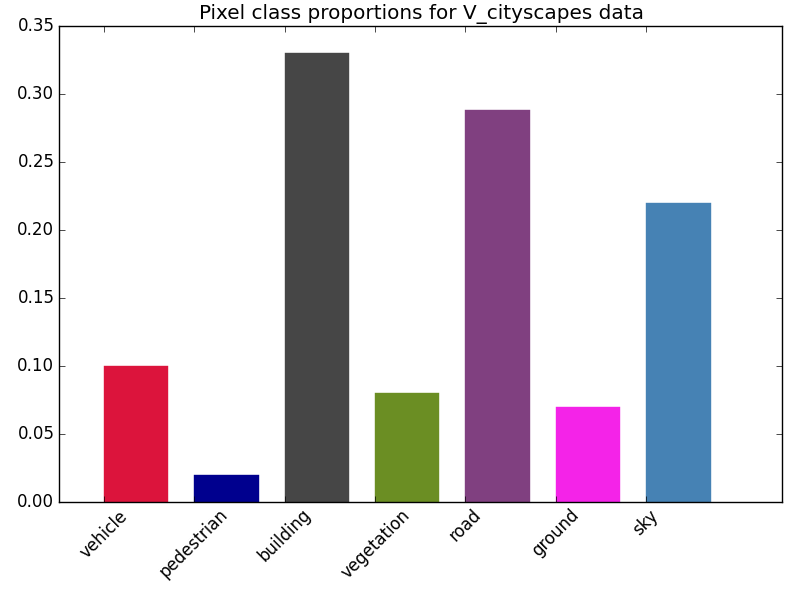}}

\rotatebox{90}{\ \ \textbf{$Camvid\ data$}}
\subfloat[Histogram of CamVid\label{fig_hist_camvid}]{\includegraphics[width=0.18\textwidth, height=2.5cm]{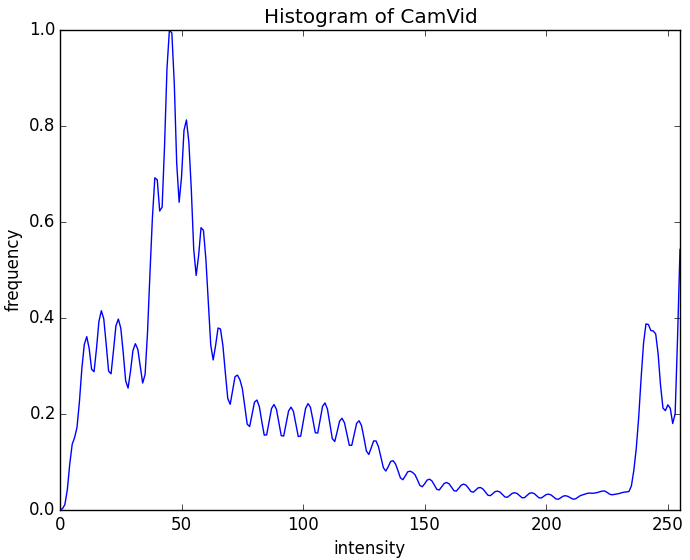}}
\subfloat[A few samples from CamVid data]{\includegraphics[width=0.18\textwidth, height=2.5cm]{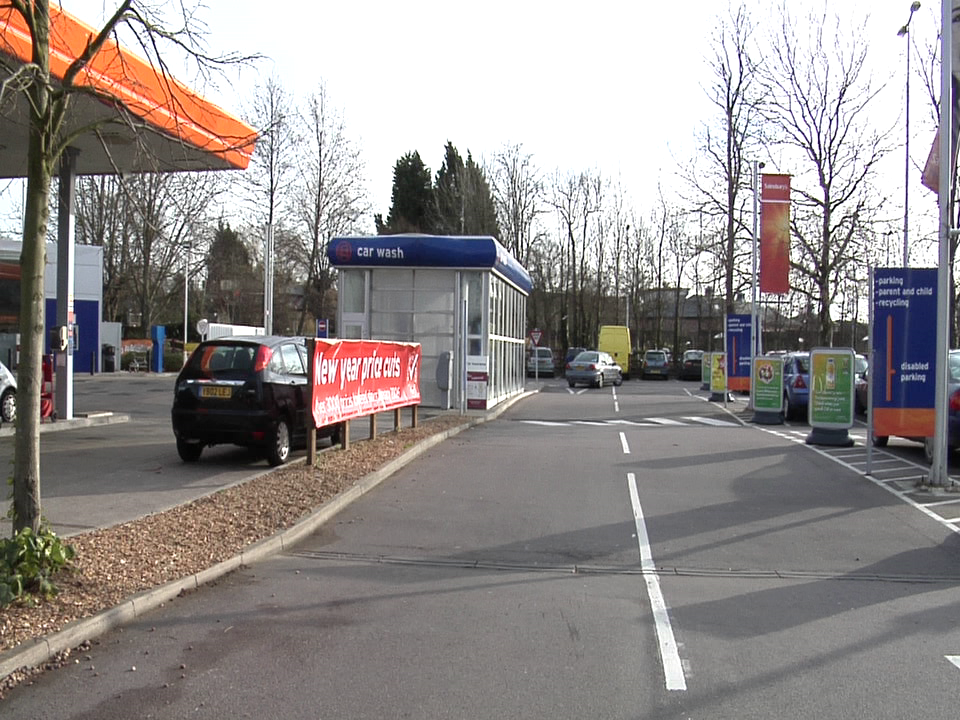}
\includegraphics[width=0.18\textwidth, height=2.5cm]{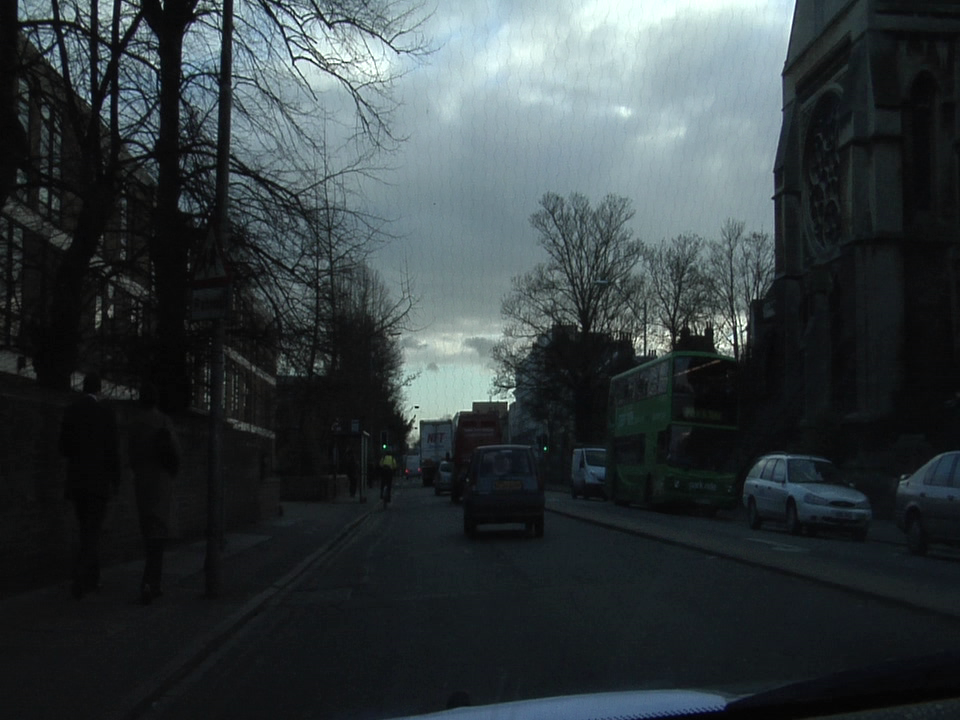}
\includegraphics[width=0.18\textwidth, height=2.5cm]{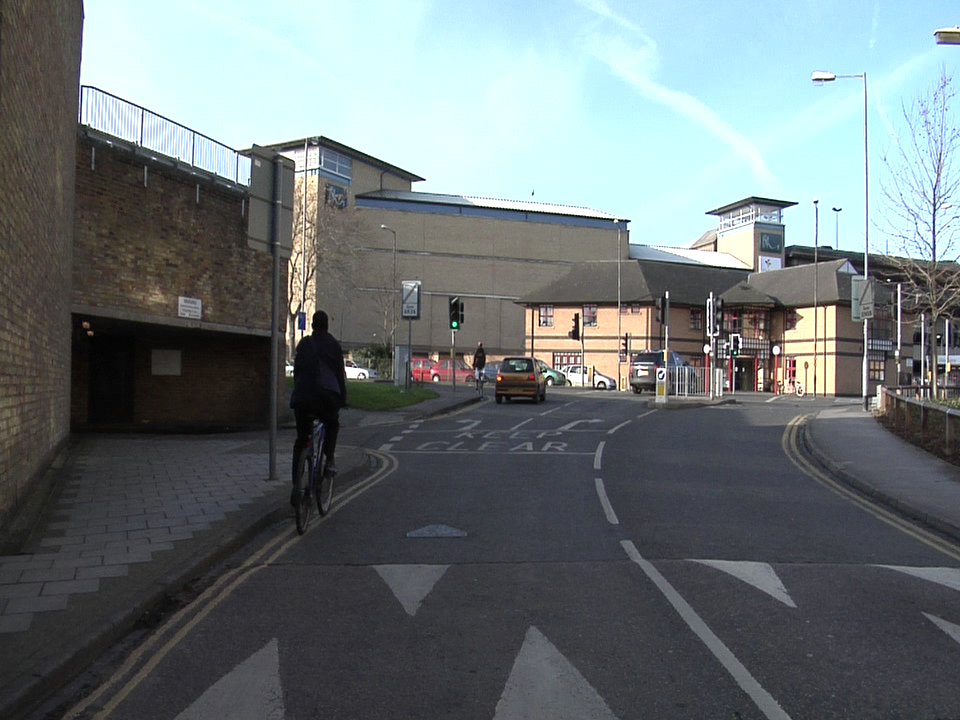}}
\subfloat[Pixel-proportions/class \label{fig_label_camvid}]{\includegraphics[width=0.18\textwidth, height=2.5cm]{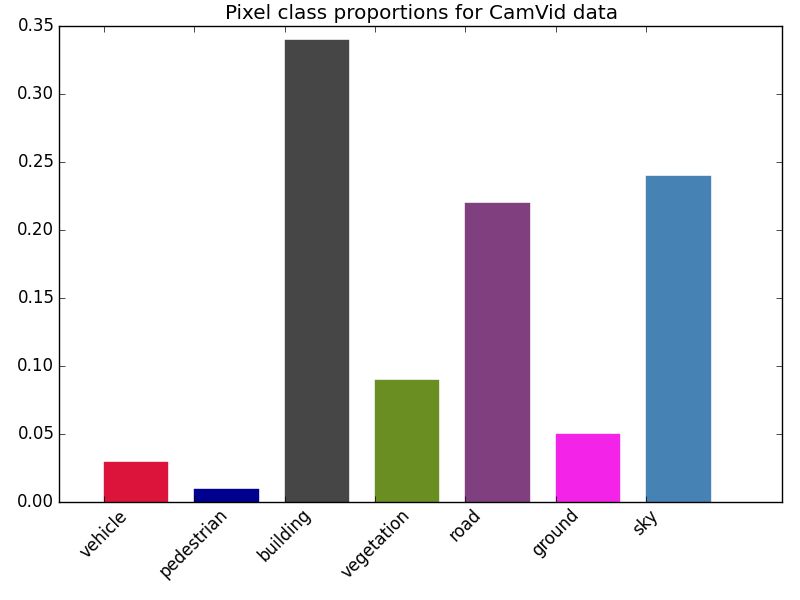}}

\rotatebox{90}{\ \ \ \ \ \textbf{$V_{camvid}$}}
\subfloat[Histogram of $V_{camvid}$\label{fig_hist_vcamvid}]{\includegraphics[width=0.18\textwidth, height=2.5cm]{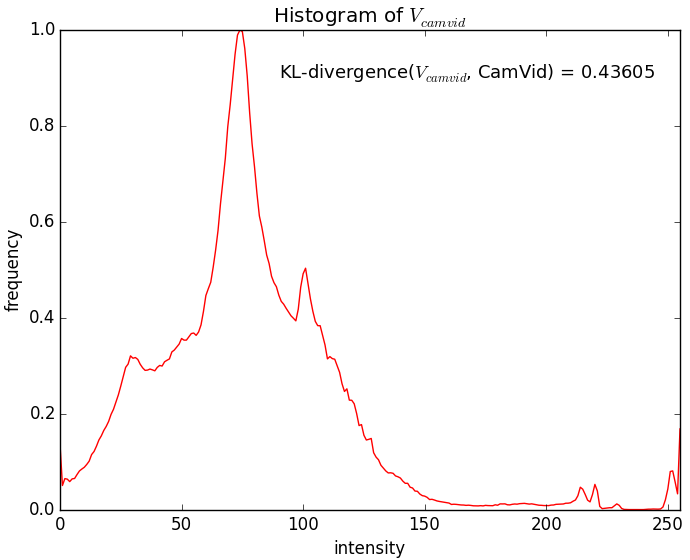}}
\subfloat[A few samples of $V_{camvid}$ sampled from the model after tuning \label{fig_vcamvid}]{\includegraphics[width=0.18\textwidth, height=2.5cm]{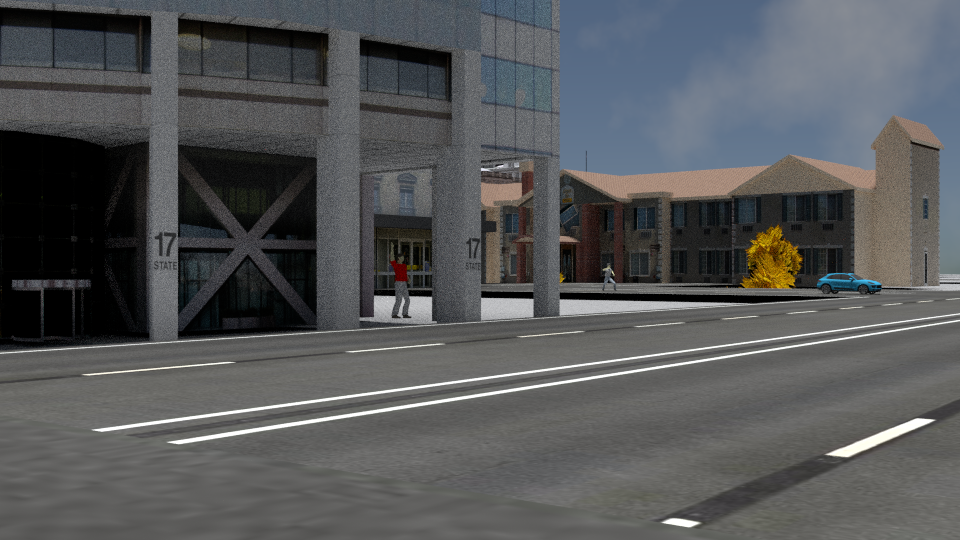}
\includegraphics[width=0.18\textwidth, height=2.5cm]{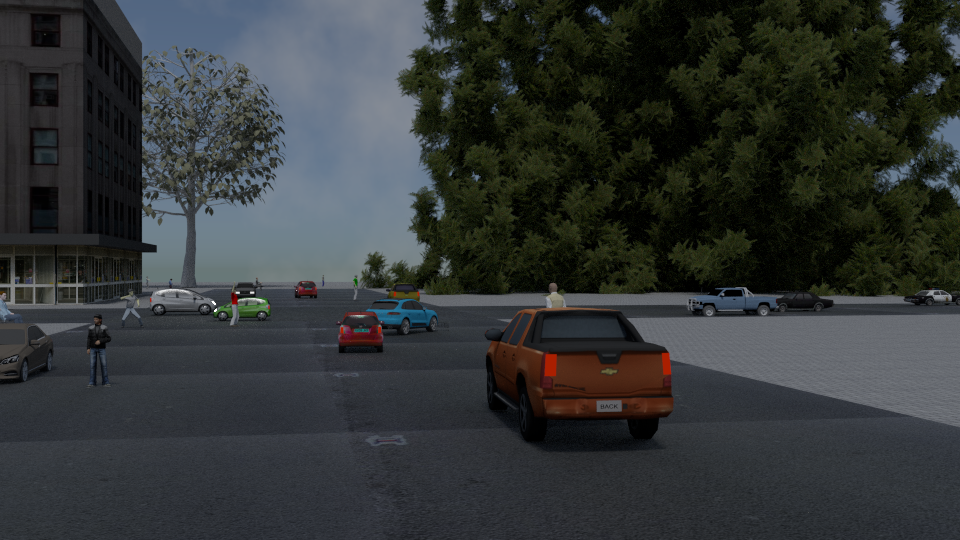}
\includegraphics[width=0.18\textwidth, height=2.5cm]{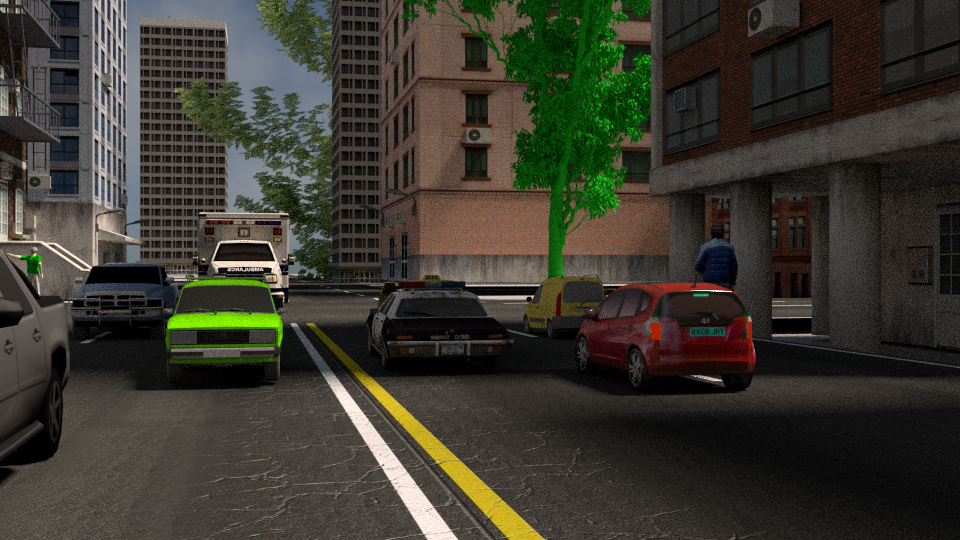}}
\subfloat[Pixel-proportions/class]{\includegraphics[width=0.18\textwidth, height=2.5cm]{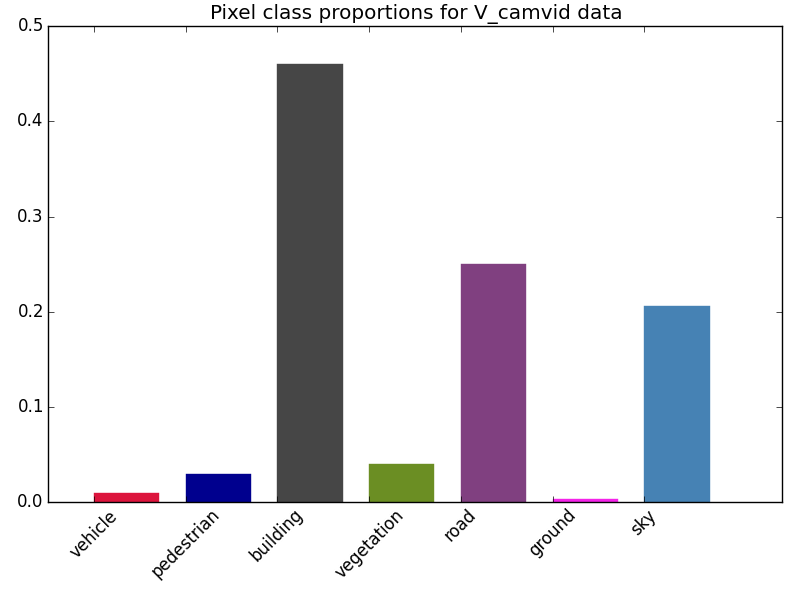}}

\caption{Qualitative comparison of training sets, both simulated and real, and their statistics before and after tuning the generative model (Best viewed in color).}
\label{fig_before_after}
\end{sidewaysfigure*}

\section{EXPERIMENTS}\label{sec_experiment}
In this section, we provide an evaluation of our generative adversarial tuning approach in terms of performance of a deep convolutional network (DCN) for urban traffic scene semantic segmentation. We choose to use a state-of-the-art DCN-based architecture as a vision system $S$ for these experiments. As we treat $S$ as a black-box, we believe that our experimental results will be of interest to other researchers using DCN-based applications. We selected two publicly available urban datasets to study the benefits of our approach for synthetic data generation.

\textbf{Vision system ($\textbf{S}$)}:
We select a state-of-the-art DCN-based architecture, i.e. DeepLab \cite{chen2014semantic} as $S$. DeepLab is an modified version of VGG-net to operate at original image resolutions, by making the following changes: 1) replacing the fully connected layers with convolutional ones, 2) skiping the last subsampling steps and up-sampling the feature-maps by using \textit{Atros} convolutions. This still results in a coarser map with a stride of 8 pixels. Hence, during training the targets, i.e. the semantic labels, are the ground truth labels subsampled by 8. During testing, bi-linear interpolation followed by a fully connected conditional random field (CRF) was used to obtain the final label maps. We modify the last layer of DeepLab from a 21-class to a 7-class,  including the categories: vehicle, pedestrian, building, vegetation, road, ground, and sky. 

\textbf{Training $\textbf{S}$}: Our DeepLab models are initialized with ImageNet pre-trained weights to avoid longer training times. Stochastic gradient descent and the cross-entropy loss function are used with an initial learning rate of 0.001, momentum of 0.9 and a weight decay of 0.0005. We use a mini-batch of 4 images and the learning rate is multiplied by 0.1 after every 2000 iterations. High-resolution input images are down-sampled by a factor 4. Training data is augmented by vertical mirror reflections and random croppings from the original resolution images, which increases the amount of data by a factor of 4. As stopping criteria, we used a fixed number of SGD iterations (30,000) in all our experiments. 
In the CRF postprocessing, we used fixed parameters in the CRF inference process (10 mean field iterations with Gaussian edge potentials as described in the \cite{chen2014semantic}) in all reported experiments. The CRF parameters are optimized on a subset of 300 images randomly selected from the training set. The peformance of DeepLab with different training-testing settings is tabulated in Table \ref{tab_results}. We report the accuracy in terms of the IoU measure for DeepLab for each of the seven classes with their average per-class and global accuracies for both real datasets we used. 

\textbf{Real world target datasets $T$}: We used CityScapes \cite{cordts2016cityscapes} and CamVid \cite{brostow2009semantic} as target datasets which are tailored for urban scene semantic segmentation. CityScapes was recorded on the streets of several European cities. It provides a diverse set of videos with public access to 3475 images with finer pixel-level annotations for semantic labels. However, in the adversarial tuning process we use 1000 randomly selected samples from CityScapes as $T$ in each iteration to train $D$ and we set $N_v=1000$ to generate 1000 samples from $P(\Theta)$. CamVid is recorded in and around the Cambridge region in UK. It provides 701 images along with high-quality semantic annotations. While tuning the generative model to CamVid, we randomly sample 500 samples from CamVid in each iteration and set $N_v=500$. 

It is worth highlighting the differences between these datasets. Each of them has been acquired in a different city or cities. The camera models used are different. Due to the geographical and demographical differences in weather, lighting, object shapes, the statistics of these dataset may differ. For instance, we computed the intensity histograms over full CityScapes and CamVid datasets, see Fig \ref{fig_hist_cityscapes} and Fig \ref{fig_hist_camvid}. For better visual comparison, we normalized the histograms with their maximum frequencies. Topologically, these histograms are quite different. Similarly, label statistics also differ, see the histograms of semantic class labels in Fig \ref{fig_label_cityscapes} and Fig \ref{fig_label_camvid}. As quantified in Table \ref{tab_results}, these statistical differences in the training datasets are reflected as performance shift of DeepLab. For instance, the DeepLab model trained on CityScapes training data ($CS\_train$) is performing at 67.71 IoU points on $CS\_val$, a validation set from CityScapes, i.e. within the same domain. This performance is reduced by nearly 13 points instead when the validation set from CamVid ($CV\_val$) is used for testing. Similar behavior is observed when transferring the DeepLab model from CityScapes to CamVid. Performance degradation when transferring from virtual to real domains is comparable. Similar observations can be found in \cite{vazquez2014virtual} in the context of pedestrian detection with a classifier based on HOG and linearSVM. 

\textbf{Virtual reality datasets ($V$)}: To quantify the performance changes due to adversarial tuning, we prepared three sets that are simulated from the initial model and the models tuned with the approach discussed in Section \ref{sec_tune} to the datasets CityScapes and CamVid. We denote them with  $V_{init}$, $V_{cityscapes}$ and $V_{camvid}$ respectively. Each set has 5000 images along with several annotations along with pixel-wise semantic labels. We first compare the performance statistics of simulated training sets against the target datasets used for adversarial tuning.  Later in the section, we also compare the generalizations of a vision system on the target dataset when it is trained on these sets separately to quantify the performance shift due to adversarially trained scene generation.

\subsection{Statistics of Training sets}
Though its difficult to appreciate significant performance changes due to adversarial training 
by visual inspection, Figures \ref{fig_vinit}, \ref{fig_vcityscapes}, and \ref{fig_vcamvid} 
can be used to obtain insights about how the training affected pixel-level labeling.
We computed histograms of pixel intensities over the full datasets $V_{init}$ generated from the initial model, our target data CityScapes and generated the the model tuned to CityScapes $V_{cityscapes}$. These plots are shown in the first column of Fig \ref{fig_before_after}. The structure of these histogram has been moved closer to the one of CityScapes through the process of tuning. Quantitatively, the KL divergence between virtual data and CityScapes data has been reduced from 0.57 before tuning to 0.44 after tuning to CityScapes. A similar behavior is observed when the model is trained on the CamVid data. Finally 
we also obtained similar histograms for the ground-truth labels. 
As with the previous comparisons, 
on can observe that the label statistics are again closer to the real datasets after tuning, as shown in the last column of Fig \ref{fig_before_after}. This evidence points to the potential usefulness of simulated datasets  as virtual proxies for these real world datasets. 

\begin{table*}[]
\small
\centering
\caption{Quantitative analysis of the performance of DeepLab models with different training-testing combinations. \\ Notation: CS and CV refers to real CityScapes and CamVid datasets respectively, and prefix 'V' represents simulated sets.}
\label{tab_results}
\begin{tabular}{@{}llllllllll@{}}
\toprule
\textit{\textbf{Training set}} & \textit{\textbf{Validation}} & \textit{\textbf{global}}             & \textit{\textbf{vehicle}} & \textit{\textbf{pedestrian}} & \textit{\textbf{building}} & \textit{\textbf{vegetation}} & \textit{\textbf{road}} & \textit{\textbf{ground}} & \textit{\textbf{sky}} \\ \midrule
\multicolumn{10}{c}{\cellcolor[HTML]{C0C0C0}\textit{Model Tuned to CityScapes data}}                                                                                                                                                                                                                    \\
V\_init                        & CS\_val                      & 49.86                                & 48                        & 53                           & 63                         & 51                           & 47                     & 34                       & 53                    \\
V\_cityscapes                  & CS\_val                      &  52.14 (\textcolor{blue}{+2.28}) & 56                        & 47                           & 65                         & 57                           & 53                     & 31                       & 56                    \\
CS\_train                      & CS\_val                      & 67.71                                & 59                        & 57                           & 73                         & 64                           & 69                     & 64                       & 88                    \\
V\_cityscapes                  & CV\_val                      & 50.28 (\textcolor{blue}{+0.43})                        & 51                        & 50                           & 55                         & 48                           & 49                     & 49                       & 50                    \\
CS\_train                      & CV\_val                      & 54.42                                & 47                        & 43                           & 55                         & 69                           & 46                     & 51                       & 70                    \\
\multicolumn{10}{c}{\cellcolor[HTML]{C0C0C0}\textit{Model Tuned to CamVid Data}}                                                                                                                                                                                                                        \\
V\_init                        & CV\_val                      & 46.42                                & 53                        & 38                           & 54                         & 35                           & 43                     & 39                       & 63                    \\
V\_camvid                      & CV\_val                      & 49.85 (\textcolor{blue}{+3.42})                        & 57                        & 34                           & 63                         & 37                           & 48                     & 44                       & 66                    \\
CV\_train                      & CV\_val                      & 67.42                                & 77                        & 34                           & 65                         & 54                           & 98                     & 45                       & 99                    \\
V\_camvid                      & CS\_val                      & 39.85 (\textcolor{red}{-6.57})                        & 35                        & 41                           & 44                         & 44                           & 32                     & 40                       & 43                    \\
CV\_train                      & CS\_val                      & 54.28                                & 46                        & 43                           & 55                         & 69                           & 46                     & 51                       & 70                    \\
\multicolumn{10}{c}{\cellcolor[HTML]{C0C0C0}\textit{Data augmentations}}                                                                                                                                                                                                                                \\
V\_init+10\%CS                 & CS\_val                      & 67.42                                & 60                        & 66                           & 52                         & 67                           & 74                     & 72                       & 81                    \\
V\_cityscapes + 10\%CS         & CS\_val                      & 70.01 (\textcolor{blue}{+2.57})                        & 68                        & 60                           & 59                         & 68                           & 77                     & 69                       & 89                    \\
V\_init+10\%CV                 & CV\_val                      & 68.85                                & 51                        & 61                           & 71                         & 67                           & 65                     & 77                       & 90                    \\
V\_camvid+10\%CV               & CV\_val                      & 70.57 (\textcolor{blue}{+1.71})                        & 63                        & 57                           & 76                         & 73                           & 67                     & 74                       & 84                    \\ \bottomrule
\end{tabular}
\end{table*}

\subsection{Generalization of DeepLab}
In our first set of experiments we used CityScapes as the target domain which means that we took the validation set from CityScapes ($CS\_{val}$) for testing. We compared the utility of simulated data generated from the initial model $V_{init}$ and the model tuned to CityScapes ($V_{cityscapes}$) in terms of generalization of the trained models to $CS\_{val}$. $V_{init}$ produced good results in classifying the objects such as building, vehicles, vegetation, roads, and sky. However, pedestrians were poorly recognized due to low frequency of occurrences and the use of low quality (low polygon meshes and textured) CAD models. However, the use of $V_{cityscapes}$, which is generated from the model tuned to real CityScapes, improved the over-all performance on global IoU by 2.28 points. This time, the per-class IoU measure on the pedestrian class also improved to some extent. This may be credited to the increased number of occurrences after tuning. This can be discerned in the bar plot of Fig \ref{fig_before_after}, last column. To measure the statistical significance of these improvements, we repeated the training-testing experiment 5 times and measured the improvement each time. The computed mean and standard deviations are $2.28 \pm 0.34$. 

In our second set of experiments we use CamVid as the target domain and take the validation set from CamVid $CV\_{val}$ for testing. We compared the utility of the simulated data generated from the initial model $V_{init}$ and the model tuned to CamVid $V_{camvid}$ in terms of the generalization from the trained models to $CV\_{val}$. $V_{init}$ already produced good results. However, the use of $V_{camvid}$ improved the overall performance, i.e. the global IoU by 3.42 points. Interestingly, the DeepLab model trained on $V_{cityscapes}$ showed improved performance also on the CamVid validation set, which however was not true the other way around 
as seen by a degradation in performance of 6.57\%. We conjecture that the high number of pedestrians and their diversity in the CityScapes set might be one of the reasons. 

In the final set of experiments, we compared the results of unsupervised adversarial tuning to those of supervised domain adaptation, i.e. augmenting the simulated data with 10\% labeled samples from the target domain. Clearly, supervised domain adaptation provides improved performance gains over our adversarial tuning approach. However, we note that our modest improvements using unsupervised learning described above were achieved without labelled samples from the target domain, thus, the costs for these improvements is low by comparison.  Instead of using the data simulated with the initial model $V_{init}$,
we improve performance on the corresponding validation sets by 2.57 and 1.71 IoU points respectively
by using data from models tuned to $V_{cityscapes}$ and $V_{camvid}$ with DeepLab. 
This 
suggests that the amount of real world labelled data required to correct for the domain-shift in order to achieve the same level of performance as $V_{init}$+10\%CS is reduced. A rough analysis using a linear fit to the empirical performance gains reported in Table \ref{tab_results} provides the observation that the amount of labelled real world data needed to reach the same level performance with  $V_{cityscapes}$ is 9\% of training data compared to the 10\% labeling of training data needed for $V_{init}$.

\section{CONCLUSIONS AND FUTURE WORK}\label{sec_conclude}
In this work, we have evaluated an adversarial approach to tune generative scene priors for the process of CG-based data generation to train CV systems. To achieve this goal, we designed a parametric scene generative model, followed by AlexNet whose output probabilities are used to update the distributions over scene parameters. Our experiments in the context of urban scene semantic segmentation with DeepLab provided evidence of improved generalization of models trained on simulated data generated from adversarially tuned scene models. These improvements were found to be on average 2.28\% and 3.42\% IoU points on two real world benchmark datasets, CityScapes and CamVid respectively. 


Our current work does not vary the intrinsic attributes of objects such as shape and texture. Instead we used a fixed set of CAD shapes and textures as a proxy to model intra-class variations. 
We expect significant performance improvements for the future when expanding the set of 3D models from the current, 
relatively small and fixed set of CAD models. A possible extension is to use component-based shape synthesis models similar to \cite{kalogerakis2012probabilistic} in order to learn distributions over object shapes. We plan to conduct more experiments to characterize the behavior of adversarial tuning by studying the variability in performance on simulated training and target domains.
Of particular interest should be relating 
the performance gains as a function of the KL-divergence between the prior distributions used for training and those of the target domains.

\section*{Acknowledgements}
This work was supported by the German Federal Ministry of Education and Research (projects 01GQ0840 and 01GQ0841) and by Continental automotive GmbH.

\newpage

{
\bibliographystyle{ieee}
\bibliography{references}
}

\end{document}